\ifdefined\pdfminorversion
\fi

\documentclass[acmtog]{acmart}

\AtBeginDocument{%
  }

\copyrightyear{2026}
\acmYear{2026}
\setcopyright{cc}
\setcctype{by}
\acmConference[SIGGRAPH Conference Papers '26]{Special Interest Group on Computer Graphics and Interactive Techniques Conference Conference Papers}{July 19--23, 2026}{Los Angeles, CA, USA}
\acmBooktitle{Special Interest Group on Computer Graphics and Interactive Techniques Conference Conference Papers (SIGGRAPH Conference Papers '26), July 19--23, 2026, Los Angeles, CA, USA}
\acmDOI{10.1145/3799902.3811149}
\acmISBN{979-8-4007-2554-8/2026/07}

\usepackage{multirow}
\usepackage{makecell}
\usepackage{pifont}

\definecolor{shadecolor}{rgb}{0.94,0.94,0.94}

\usepackage{amsmath,amsfonts,bm}

\def\eqref#1{equation~\ref{#1}}

\def\1{\bm{1}}

\DeclareMathAlphabet{\mathsfit}{\encodingdefault}{\sfdefault}{m}{sl}
\SetMathAlphabet{\mathsfit}{bold}{\encodingdefault}{\sfdefault}{bx}{n}

\newcommand{\model}{SCOPE}

\begin{document}

\title[\model]{\model: Scale-Consistent One-Pass Estimation of 3D Geometry}

\author{Zheng Zhang}
\orcid{0009-0001-5282-3661}
\affiliation{%
  \institution{The University of Hong Kong}
  \city{Hong Kong}
  \country{Hong Kong}
}
\email{u3009782@connect.hku.hk}

\author{Lihe Yang}
\orcid{0009-0007-8600-9733}
\affiliation{%
  \institution{The University of Hong Kong}
  \city{Hong Kong}
  \country{Hong Kong}
}
\email{lihe.yang.cs@gmail.com}

\author{Tianyu Yang}
\orcid{0000-0002-9674-5220}
\affiliation{%
  \institution{Alibaba Group}
  \city{Hong Kong}
  \country{Hong Kong}
}
\email{tianyu-yang@outlook.com}

\author{Chaohui Yu}
\orcid{0000-0002-7852-4491}
\affiliation{%
  \institution{Alibaba Group}
  \city{Hangzhou}
  \country{China}
}
\email{huakun.ych@alibaba-inc.com}

\author{Yixing Lao}
\orcid{0000-0001-8338-3577}
\affiliation{%
  \institution{The University of Hong Kong}
  \city{Hong Kong}
  \country{Hong Kong}
}
\email{yixing.lao@gmail.com}

\author{Xiaoyang Guo}
\orcid{0000-0001-8265-7441}
\affiliation{%
  \institution{Horizon Robotics}
  \city{Shanghai}
  \country{China}
}
\email{xiaoyang.guo1995@gmail.com}

\author{Biao Gong}
\orcid{0000-0002-6156-0816}
\affiliation{%
  \institution{Ant Group}
  \city{Hangzhou}
  \country{China}
}
\email{a.biao.gong@gmail.com}

\author{Fan Wang}
\orcid{0000-0001-7320-1119}
\affiliation{%
  \institution{Alibaba Group}
  \city{Hangzhou}
  \country{China}
}
\email{fan.w@alibaba-inc.com}

\author{Hengshuang Zhao}
\orcid{0000-0001-8277-2706}
\affiliation{%
  \institution{The University of Hong Kong}
  \city{Hong Kong}
  \country{Hong Kong}
}
\email{hszhao@cs.hku.hk}

\renewcommand{\shortauthors}{Zhang et al.}

\makeatletter
\def\@typeset@author@line{%
  \andify\@currentauthors\par\noindent
  \@currentauthors\gdef\@currentauthors{}%
  \ifx\@currentaffiliations\@empty\else
    \andify\@currentaffiliations
      \unskip, {\@currentaffiliations}\par
  \fi
  \gdef\@currentaffiliations{}}
\makeatother

\begin{abstract}

    We present \model{} (Scale-Consistent One-Pass Estimation of 3D Geometry), a novel approach for estimating 3D geometry from extended monocular video sequences, where existing methods struggle to maintain both geometric accuracy and temporal consistency across hundreds of frames. Our approach generates affine-invariant 3D point maps with shared parameters across entire sequences, enabling consistent scale-invariant representations. We introduce three key innovations: viewpoint-invariant geometry aligning multi-perspective points in a unified reference frame; appearance-invariant learning enforcing consistency across exponential timescales; and frequency-modulated positioning enabling extrapolation to sequences vastly exceeding training length. Experiments across diverse datasets demonstrate significant improvements, reducing relative point map error by 24.2\% and temporal alignment error by 34.9\% on ScanNet compared to state-of-the-art methods. Our approach handles challenging scenarios with complex camera trajectories and lighting variations while efficiently processing extended sequences in a single pass. Project page: \url{https://scope3d.github.io/}.

\end{abstract}


\begin{teaserfigure}
  \centering
  \includegraphics[width=\textwidth]{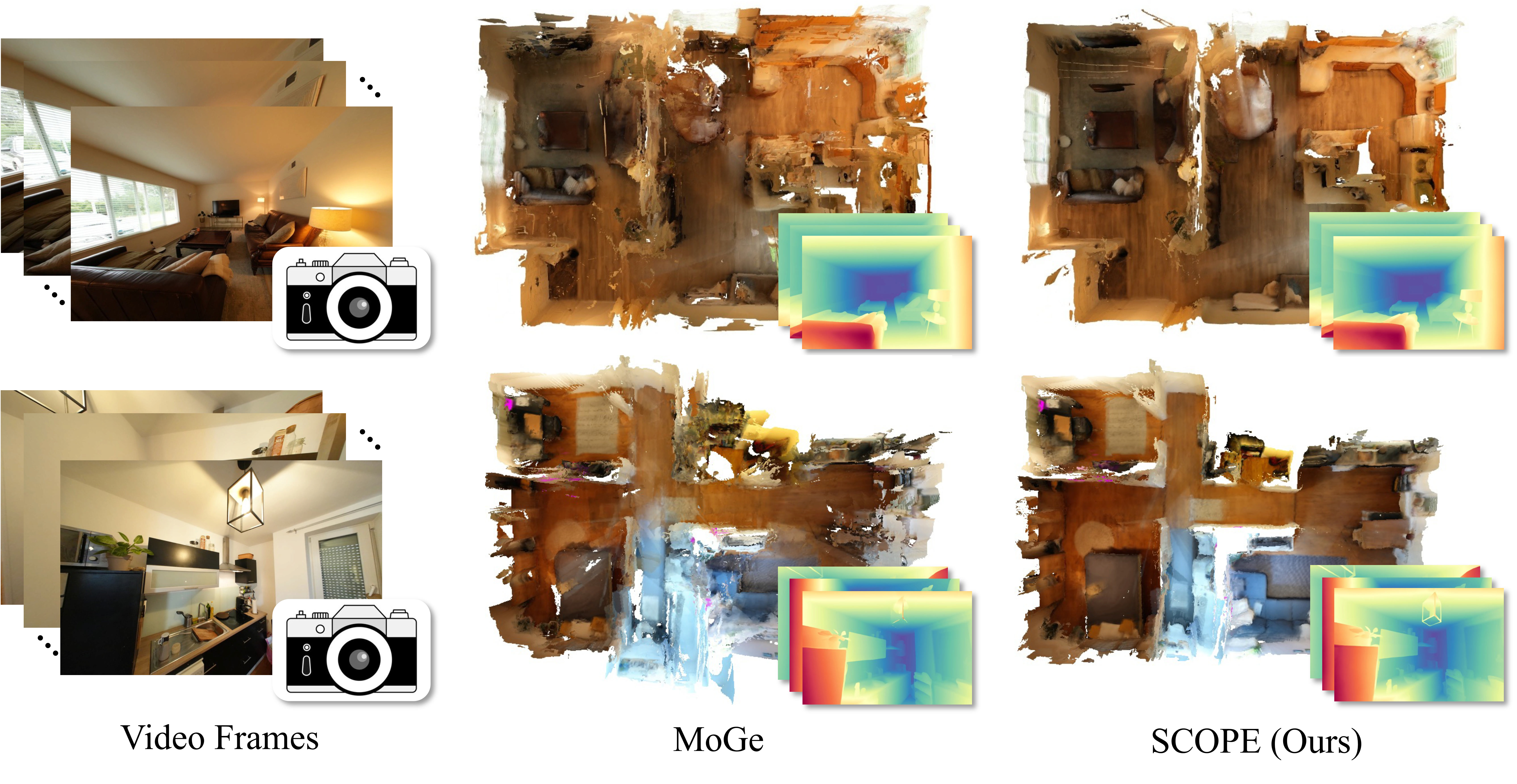}
  \caption{
    Given a sequence of video frames, \model{} is capable of predicting scale-invariant and temporal-consistent point maps in a single forward pass. We visualize the 3D mesh reconstructed by TSDF integration of 100 point maps predicted by \model{} in a single shot, in comparison with MoGe~\citep{moge}, using input frames from the ScanNet++ dataset~\citep{yeshwanth2023scannetpp}. \model{} maintains geometric accuracy and long-range consistency across hundreds of frames with minimum drift, enabling high-quality 3D reconstruction.
  }
  \Description{Teaser figure comparing 3D reconstruction quality from monocular video: input frames, a baseline method, and \model.}
  \label{fig:teaser}
\end{teaserfigure}

\maketitle

\section{Introduction}

Estimating 3D geometry from monocular videos is a fundamental challenge in computer vision with diverse applications in novel view synthesis, autonomous navigation, virtual reality, and 3D/4D reconstruction. Despite significant advances in single-image depth estimation, video-based approaches still struggle with two critical challenges: achieving high geometric accuracy at multiple scales within each frame and across the global coordinate system, while maintaining temporal consistency throughout sequences of hundreds of frames without scale drift.

Existing methods typically excel in one area at the expense of the other. Single-image approaches like MoGe~\citep{moge} capture detailed geometry but produce inconsistent results when applied frame-by-frame to videos. Conversely, video-specific methods~\citep{yang2024depthanyvideo,hu2025-DepthCrafter,video_depth_anything,dust3r,monst3r} inherently lack geometric precision while providing only short-term consistency, still exhibiting significant scale drift in longer sequences. Traditional approaches rely on optical flow constraints~\citep{NVDS,NVDSPLUS} that only link adjacent frames, failing to prevent accumulation of errors. Video diffusion models offer consistency through learned priors but at significant computational cost. Recent transformer-based approaches like VGGT~\citep{wang2025vggt} can process longer sequences but lack effective temporal position encoding, limiting their effectiveness with complex camera motions.

Processing lengthy sequences with geometric and temporal accuracy requires simultaneous consideration of hundreds of frames with precise temporal position encoding to handle complex scene transformations. However, memory constraints make training on such long sequences impractical. This creates a fundamental tension: models need to extrapolate effectively to sequences far longer than their training examples. With robust extrapolation capabilities, overlapping inference techniques can achieve minimal drift by maintaining substantial frame overlap between consecutive windows, effectively scaling to unlimited sequence lengths.

We present a novel approach generating affine-invariant 3D point maps from RGB videos with both geometric precision and long-range temporal consistency. Our method produces point maps where all frames share the same scale and shift parameters, with a unified optimization approach to recover scale-invariant representations for downstream applications. Our key innovations include viewpoint-invariant geometry transforming points from multiple perspectives into a shared reference frame through camera pose integration; appearance-invariant learning that supervises geometric consistency across exponential time scales while isolating persistent structural features from transient visual conditions; and adaptive frequency-modulated positioning implementing an NTK-guided rotary scheme with strategic training-time extrapolation simulation to process sequences orders of magnitude longer than training examples. We compare with metric-depth methods such as MoGe-2~\citep{moge2}, VGGT~\citep{wang2025vggt} and its derived $\pi^3$ model~\citep{pi3}, and DUSt3R-derived multiview methods~\citep{dust3r,mast3rslam,monst3r}, and report component ablations covering the architecture, cross-frame loss, temporal regularization, appearance augmentation, and positional extrapolation. Our approach significantly outperforms previous methods, reducing relative point map error by 24.2\% on ScanNet and temporal alignment error by 34.9\% compared to existing approaches, while maintaining superior performance across diverse datasets from synthetic animations to real-world driving scenarios.

\section{Related Work}
\noindent\paragraph{Monocular depth estimation.}
Recent advances in monocular depth estimation have significantly improved both geometric accuracy and generalization. Early supervised approaches~\citep{eigen2014depth,fu2018deep,adabins,zoedepth} were limited by domain-specific datasets. More recent methods overcame this limitation through affine-invariant representations~\citep{midas,midasv31,dpt} or scale alignment techniques~\citep{metric3d,metric3dv2}. Large-scale data-driven approaches~\citep{depth_anything_v1,depth_anything_v2} and diffusion-based models~\citep{marigold,depthfm,geowizard} have further enhanced generalization to diverse scenarios. While some methods~\citep{yin2021learning,unidepth,unidepthv2,depthpro} predict both depth and camera intrinsics, they often lack precision in local geometry. MoGe~\citep{moge} achieves superior geometric accuracy through multi-scale supervision but operates only on single images, lacking cross-frame consistency.

\noindent\paragraph{Video-based depth estimation.}
Extending depth estimation to video introduces temporal consistency challenges. Video diffusion models~\citep{hu2025-DepthCrafter,yang2024depthanyvideo} provide inherent coherence but at high computational cost. For sequences longer than training examples, strategies include sliding windows~\citep{hu2025-DepthCrafter,video_depth_anything}, keyframe conditioning~\citep{yang2024depthanyvideo}, and global attention~\citep{wang2025vggt}. However, these methods still exhibit scale drift over extended sequences or struggle with complex camera trajectories. Current approaches typically excel at either geometric accuracy or temporal consistency, rarely achieving both across hundreds of frames.

\noindent\paragraph{Video and multiview geometry.}
Single-frame point-map estimators such as DepthPro~\citep{depthpro}, MoGe~\citep{moge}, and MoGe-2~\citep{moge2} recover detailed per-frame geometry but do not impose long-range sequence-level consistency. Multiview systems such as DUSt3R~\citep{dust3r}, MonST3R~\citep{monst3r}, MASt3R-SLAM~\citep{mast3rslam}, VGGT~\citep{wang2025vggt}, and $\pi^3$~\citep{pi3} reason jointly over multiple images and produce globally organized geometry or camera motion in a single coordinate frame. Their primary objective is multiview reconstruction rather than long-range temporal stability across hundreds of frames, so they do not specifically enforce a shared per-sequence scale or temporal regularity over extended monocular videos. \model{} predicts per-frame point maps in each camera coordinate system while enforcing a shared sequence scale and temporal stability over hundreds of frames.

\noindent\paragraph{Positional encoding for extrapolation.}
Transformers struggle with sequences longer than their training examples. While standard sinusoidal encodings~\citep{vaswani2017attention} and learned embeddings have limited extrapolation capabilities, Rotary Position Encoding (RoPE)~\citep{su2021roformer} better generalizes by encoding relative positions through complex plane rotations. Strategic frequency adjustments~\citep{chen2023extending,peng2024yarn} and NTK-aware adaptations~\citep{peng2023ntk,sun2022length} preserve both local details and global structure during extrapolation. Our work adapts these techniques, primarily developed for language models, to video-based 3D reconstruction, enabling effective processing of sequences substantially longer than training examples.

\section{Method}

We present a novel approach for generating geometrically accurate and temporally consistent 3D point maps from RGB videos. Our method addresses two critical challenges: producing geometrically precise representations for each frame, and maintaining long-range temporal consistency across hundreds of frames - essential requirements for downstream 3D reconstruction tasks.

\subsection{Geometry-Aware Video Point Map Estimation}

\paragraph{Task definition.}
Given an RGB video sequence $\mathbf{I} = \{I_1, I_2, ..., I_T\}$ with $T$ frames, our goal is to estimate scale-invariant and temporally consistent 3D point maps from unposed monocular videos. Specifically, we predict a sequence of 3D point maps $\mathbf{P} = \{P_1, P_2, ..., P_T\}$, where $P_t \in \mathbb{R}^{H \times W \times 3}$ represents the 3D coordinates of each pixel in frame $t$ within that frame's camera coordinate system. \textbf{Training setup:} During training, our method takes multi-frame RGB images as network input and optionally uses ground truth camera poses solely for computing the cross-frame geometric loss. The poses enable multi-scale geometric supervision by transforming predicted point clouds to a common reference frame, but are not required for all training data. \textbf{Inference setup:} At inference, our method requires only multi-frame RGB images as input and outputs scale-consistent point maps in each frame's camera coordinate system. These point maps can subsequently serve as input to methods like MegaSAM~\citep{megasam} to estimate camera parameters and enable high-quality 4D reconstruction.
Although each $P_t$ is expressed in its own camera coordinate system, a point map is not equivalent to predicting depth alone: it also encodes per-pixel ray directions, which lets us recover one shared focal length and shift for the whole sequence. All frames then share the same affine ambiguity, whereas independently predicted depth maps can fit each frame with different scale and shift parameters.

\begin{sloppypar}
\paragraph{Scale-invariant representation.}\label{subsec:scale-invariant}
Our model predicts affine-invariant point maps following MoGe~\citep{moge}, where each point map is agnostic to global scale $s$ and offset $\mathbf{t}$. The key distinction is that our video sequence shares these parameters: $P_i \cong s P_i + \mathbf{t}$, $\forall i \in [1, T]$. During inference, we recover a shared focal length $f$ and Z-axis shift $t_z$ by minimizing the projection error:
\end{sloppypar}

\begin{equation}
\min_{f, t_z} \sum_{t=1}^{T}\sum_{i=1}^{N} \left(\frac{f x_{t,i}}{z_{t,i} + t_z} - u_{t,i} \right)^2 + \left(\frac{f y_{t,i}}{z_{t,i} + t_z} - v_{t,i} \right)^2,
\end{equation}

\noindent where $(x_{t,i}, y_{t,i}, z_{t,i})$ are the predicted 3D coordinates and $(u_{t,i}, v_{t,i})$ are the corresponding 2D pixel coordinates. This ensures a metrically consistent representation across the entire video, essential for 3D reconstruction tasks. By recovering the shift parameter, we transform our predictions into scale-invariant point maps, making them directly applicable for downstream tasks such as 3D reconstruction and novel view synthesis.

\paragraph{Geometric accuracy through multi-scale training.} We retain MoGe's affine-invariant global alignment, multi-scale local geometry supervision, and normal consistency losses as the per-frame geometric foundation, then extend them to video with cross-frame constraints and temporal supervision. We also include spatial gradient regularization~\citep{depthpro,depth_anything_v2,yin2021virtual,Yin2019enforcing} to preserve local surface detail.

\begin{figure*}[t!]
\centering
\includegraphics[width=\textwidth]{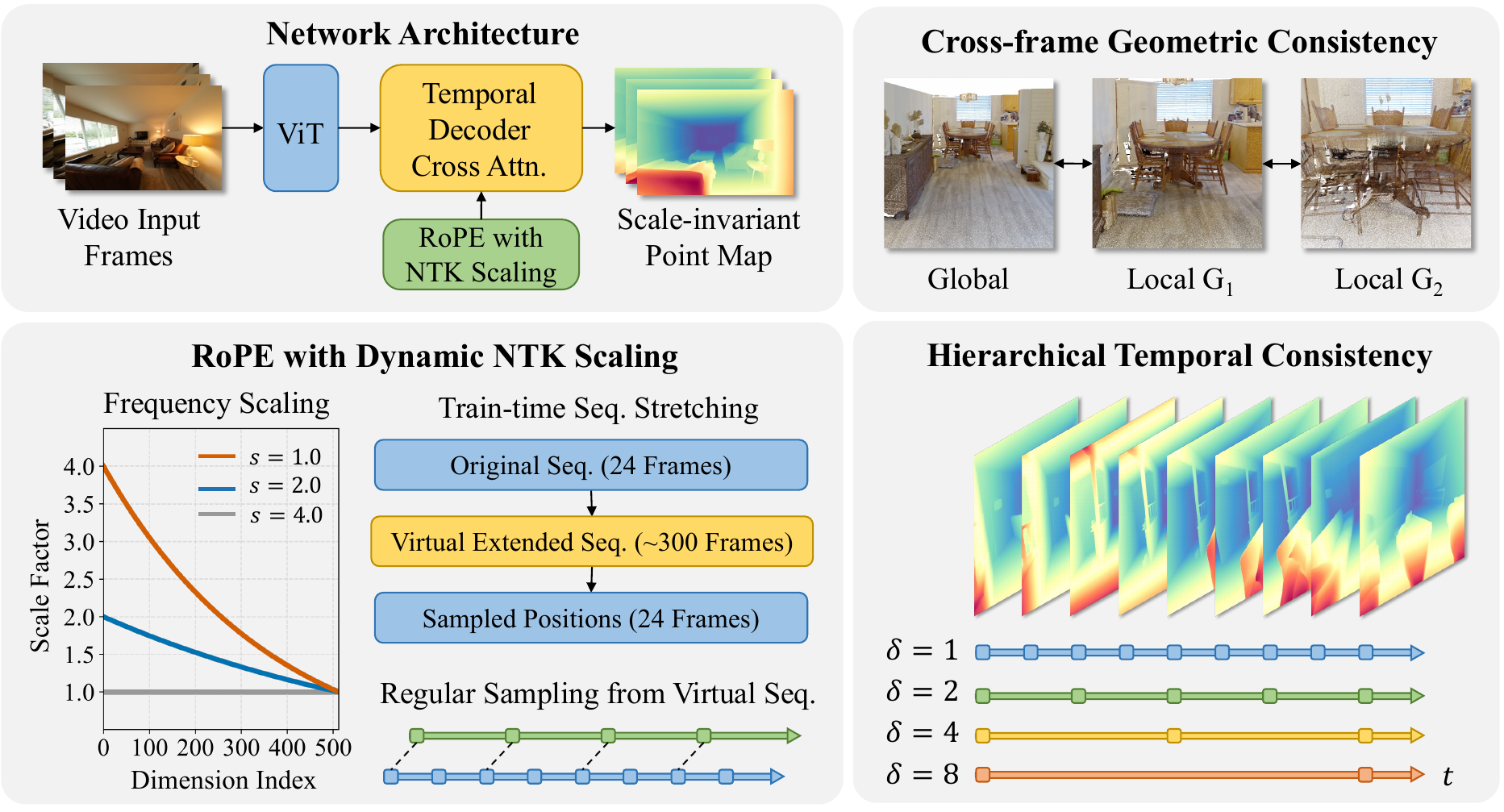}
\caption{
  \textbf{Overview of \model.~}
  \textit{Top-Left:} \model{} consists of a ViT backbone that processes video input frames, followed by a temporal decoder with cross-attention and dynamic NTK scaling RoPE, producing scale-invariant point maps (Sec.~\ref{subsec:scale-invariant}).
  \textit{Top-Right:} Cross-frame geometric consistency enforced across global and local geometric levels ($G_1$, $G_2$) to maintain structural coherence across frames (Sec.~\ref{subsec:cross-frame}).
  \textit{Bottom-Left:} RoPE with dynamic NTK scaling applied to extend sequence context, using frequency scaling that adaptively weights dimensions based on scale factor, and train-time sequence stretching that creates a virtual extended sequence to sample positions (Sec.~\ref{subsec:freq-modulated}).
  \textit{Bottom-Right:} Hierarchical temporal consistency constraints applied multiple temporal strides ($\delta = 1, 2, 4, 8$) to enforce smooth, consistent point map predictions across time (Sec.~\ref{subsec:structure-preserving}).
}
\Description{Overview diagram of the \model{} architecture, illustrating the temporal decoder, cross-frame geometric consistency, and hierarchical temporal consistency with NTK-scaled RoPE.}
\label{fig:method}
\end{figure*}

\paragraph{Cross-frame geometric constraints.}\label{subsec:cross-frame}
To enforce geometric consistency across frames, we transform all points to a common reference frame using camera poses. We randomly select one frame from the sequence as the reference frame for each training iteration, providing diverse viewpoints during training. This process involves first converting points from individual camera coordinates to world coordinates using the camera-to-world transforms, and then transforming these world points to the randomly selected reference frame. This transformation allows us to directly compare geometric structures captured from different viewpoints within a unified coordinate system.

We then apply a multi-scale geometric loss framework to the points in this common reference frame:

\begin{equation}
\begin{split}
\mathcal{L}_{cross} = \sum_{l \in \{1, G_1, G_2\}} \frac{1}{|C_l|} \sum_{c \in C_l} \frac{1}{|M_c|} \sum_{i \in M_c} w_i \\
\cdot \left\|s_c \mathbf{p}_{pred}^{ref}[i] + \mathbf{t}_c - \mathbf{p}_{gt}^{ref}[i]\right\|_1,
\end{split}
\end{equation}

\noindent where $l$ is the grid size (with $l=1$ representing global alignment), $C_l$ is the set of cells at grid size $l$, $M_c$ is the set of valid points in cell $c$, $w_i=\operatorname{clip}(1/|z_i|)$ is a depth-aware weight that balances near and far geometry without allowing extremely small depths to dominate, and $(s_c, \mathbf{t}_c)$ are alignment parameters computed independently for each cell.

For global alignment ($l=1$), the entire point cloud is treated as a single cell. For local alignment, we divide the 3D space into a grid of $G_l \times G_l \times G_l$ cells. Our implementation uses grid sizes of 4 and 16, allowing the model to capture both coarse structure and fine details across the entire temporal sequence. The reference frame is randomly sampled at each training iteration so the loss is not tied to one fixed camera view, and the same geometry is supervised under different reference choices. By enforcing geometric consistency at multiple scales, our approach ensures that the predicted point maps maintain both local detail and global structure across the video.

\subsection{Long-Range Temporal Consistency}

\paragraph{Temporal consistency challenges.} Downstream reconstruction tasks require point maps that exhibit: (1) consistent geometric accuracy both within individual frames and across the entire sequence at local and global scales, and (2) temporal stability over extended sequences rather than just between adjacent frames. When these requirements aren't met, particularly under challenging conditions with dramatic lighting changes or significant camera movements, scale drift can accumulate, severely degrading reconstruction quality and producing distorted or fragmented results.

Recent video diffusion model-based approaches~\citep{hu2025-DepthCrafter, yang2024depthanyvideo, shao2024learningtemporallyconsistentvideo} leverage inherent temporal consistency mechanisms, but suffer from significant computational inefficiency. Other methods utilize optical flow-based losses~\citep{NVDS, NVDSPLUS,kuang2024bufferanytimezeroshotvideo, video_depth_anything} to maintain consistency between adjacent frames. However, these approaches only constrain relationships between consecutive frames, causing error accumulation over longer sequences. Furthermore, they struggle with large camera motions, which can substantially degrade depth prediction accuracy by introducing conflicting geometric constraints when camera viewpoint changes significantly.

\paragraph{Structure-Preserving temporal supervision.}\label{subsec:structure-preserving} 
To address fundamental limitations in temporal consistency, we introduce a hierarchical derivative supervision framework that operates across multiple time scales:

\begin{equation}
\mathcal{L}_{temp} = \sum_{s=0}^{S-1} \frac{1}{|M_s|} \sum_{t=1}^{T-\delta_s} \sum_{i \in \mathcal{M}_{t,t+\delta_s}} w_{t,i} \cdot \left|\frac{\partial D_{pred}}{\partial t}(t,i) - \frac{\partial D_{gt}}{\partial t}(t,i)\right|,
\end{equation}

\noindent where $s$ indexes temporal scale, $S$ is the total number of scales, $\delta_s = 2^s$ represents exponentially increasing time intervals, $T$ is the sequence length, $\mathcal{M}_{t,t+\delta_s}$ denotes valid corresponding pixels between frames $t$ and $t+\delta_s$, $|M_s|$ is the total number of valid pixels at scale $s$, $w_{t,i}$ uses the same clipped inverse-depth form as above, and $\frac{\partial D}{\partial t}$ represents the temporal derivative of depth values. This first-order constraint complements zero-th order depth supervision: after per-frame affine alignment, residual depth errors can drift slowly across frames, and the derivative term penalizes such drift directly at multiple time scales.

To disentangle geometric structure from appearance variations, we apply frame-specific augmentations with independently sampled color transformations and blur patterns across the sequence. This forces the model to focus on invariant geometric features while ignoring transient visual cues, enabling robust geometric consistency even under dramatic lighting changes and complex camera movements that typically challenge conventional methods.

\paragraph{Scaling beyond memory constraints.} Processing hundreds of frames simultaneously during training is infeasible due to memory constraints. Existing methods address this limitation in various ways: DepthCrafter~\citep{hu2025-DepthCrafter} and Video Depth Anything~\citep{video_depth_anything} use overlapping frame windows during inference, but suffer from scale drift due to limited training sequence length. Depth Anything Video~\citep{yang2024depthanyvideo} processes key frames first and then uses them as conditions for other frames, but this approach has limited scalability and reduced efficiency. VGGT~\citep{wang2025vggt} employs global attention without temporal information injection, struggling with complex camera trajectories where temporal relationships are critical.

\paragraph{Frequency-modulated extrapolation.}\label{subsec:freq-modulated}
To ensure robust handling of complex spatial relationships while enabling effective extrapolation to sequences much longer than those seen during training, we employ a specialized Rotary Position Encoding (RoPE)~\citep{su2021roformer,chen2023extending,peng2023ntk,sun2022length} with Neural Tangent Kernel (NTK) adaptation.

Our implementation computes frequency components with dynamic NTK scaling:
\begin{equation}
\theta_{i,j} = \frac{j \cdot s^{(1-\frac{i}{d})}}{10000^{\frac{2i}{d}}},
\end{equation}

\noindent where $\theta_{i,j}$ is the rotation angle, $j$ is the position index, $i$ indexes the frequency dimension, $d$ is the embedding dimension, and $s = \frac{L_{seq}}{L_{train}}$ is a scaling factor applied when inference sequence length exceeds training length. This adaptive scaling preserves the model's capacity to capture both fine-grained temporal patterns and global structure by applying graduated adjustments across the frequency spectrum—attenuating changes to high-frequency components that encode local details while amplifying adjustments to low-frequency components that capture long-range dependencies.

During training, we randomly apply sequence stretching with 50\% probability, where we generate position encodings for a virtual extended sequence and sample them at appropriate intervals to match the original sequence length. Mathematically, this involves computing $\theta_{i,j}'$ for a virtual sequence of length $L_{virtual} = L_{seq} \cdot r$ (where $r$ is randomly sampled) and then sampling positions $j' = j \cdot r$ to obtain the final encodings. This technique simulates extrapolation during training, teaching the model to handle sequences significantly longer than those in the training data. Stretching modifies only the positional indices; the RGB frames, frame order, and motion are unchanged.

To further enhance temporal generalization, we employ variable temporal context windows during training. While maintaining a fixed 24-frame input size, we dynamically adjust the temporal stride between frames, allowing these 24 frames to represent contexts spanning from densely sampled short sequences to sparsely sampled long sequences of several hundred frames. This adaptive sampling strategy complements our position encoding approach, enabling the model to simultaneously learn representations for both fine-grained frame-to-frame transitions and long-range temporal relationships.

\section{Experiment}

\subsection{Implementation Details}

\paragraph{Model architecture.} Our model builds upon MoGe~\citep{moge} by integrating temporal modeling capabilities through strategically placed temporal attention modules in the decoder. Specifically, we insert one transformer block after each of the four decoder feature levels; each block uses 8 attention heads and adds 7.68M parameters in total. The blocks are inserted only in the decoder, leaving the MoGe image encoder unchanged. We employ DINOv2-L~\citep{dinov2} as our visual encoder and initialize all parameters from MoGe's pretrained weights.

\begin{sloppypar}
\paragraph{Training datasets.} We train on synthetic datasets including TartanAir~\citep{tartanair}, PointOdyssey~\citep{pointodyssey}, SPRING~\citep{spring}, VKitti2~\citep{vkitti}, Lightwheel~\citep{lightwheel}, Hypersim~\citep{hypersim}, GTAIM~\citep{gtaim}, MVSSynth~\citep{mvsynth}, US4K~\citep{us4k}, GTASFM~\citep{gtasfm}, IRS~\citep{irs}, MidAir~\citep{midair}.
We use ground-truth camera poses only from synthetic training sequences that provide camera trajectories to compute $\mathcal{L}_{cross}$; $\mathcal{L}_{temp}$ is pose-free and can be applied whenever valid temporal correspondences and depth supervision are available.
\end{sloppypar} 

\paragraph{Optimization strategy.} We optimize using AdamW following MoGe's base configuration with learning rates of $10^{-4}$ for decoder and $10^{-5}$ for encoder parameters. These rates are dynamically scaled according to the square root of batch size ratio, using a reference batch size of 32 frames as baseline. Our learning schedule employs warmup, linear decay, and step decay phases with all milestone parameters proportionally adjusted based on total iteration count. Throughout training, we preserve input aspect ratios while resizing images to maintain spatial relationships.

\paragraph{Loss functions.} Our loss function integrates MoGe's original components with additional spatial and temporal consistency objectives. We maintain the affine-invariant global loss (weight 1.0), multi-scale local losses at levels 4, 16, and 64 (weights 1.0 each), normal loss (1.0), and mask loss (1.0). We adopt established spatial gradient loss (4.0) to preserve depth details, and introduce our proposed $\mathcal{L}_{temp}$ (2.0) and $\mathcal{L}_{cross}$ (1.0). For frame-specific augmentation, we apply color jitter and Gaussian blur with 0.5 probability to enhance robustness to appearance variations.

\paragraph{Computational resources.} We trained our final model on 16 NVIDIA H20 GPUs for approximately 4.3 days. Each ablation study experiment required approximately 0.6 days of training on the same hardware configuration.

\begin{table*}[t]
\caption{
    \textbf{Evaluation on point map estimation and depth estimation.} 
    Results are aligned with the ground truth by optimizing a shared scale factor across the entire video.
    Lower values are better for $\text{Rel}$ and $\text{TAE}$ ($\downarrow$), while higher values are better for $\delta$ ($\uparrow$).
    The best results in each column are highlighted in \textbf{bold}. 
    \textcolor{gray}{Gray} values indicate methods trained on ScanNet.
}
\label{tab:comp_point}
\centering
\resizebox{\textwidth}{!}{%
\begin{tabular}{l|cc|ccc|cc|cc|c}
\toprule
\multirow{2}{*}{Method} & \multicolumn{2}{c|}{Sintel} & \multicolumn{3}{c|}{ScanNet} & \multicolumn{2}{c|}{Bonn} & \multicolumn{2}{c|}{KITTI} & Avg. \\
& $\text{Rel}^{p}\!\downarrow$ & $\delta^{p}\!\uparrow$ & $\text{Rel}^{p}\!\downarrow$ & $\delta^{p}\!\uparrow$ & $\text{TAE}^{p}\!\downarrow$ & $\text{Rel}^{p}\!\downarrow$ & $\delta^{p}\!\uparrow$ & $\text{Rel}^{p}\!\downarrow$ & $\delta^{p}\!\uparrow$ & Rank $\!\downarrow$ \\
\midrule
DepthPro~\citep{depthpro} & 0.400 & 0.441 & \textcolor{gray}{0.132} & \textcolor{gray}{0.942} & \textcolor{gray}{0.095} & 0.130 & 0.975 & 0.191 & 0.810 & 3.67 \\
VGGT~\citep{wang2025vggt} & 0.382 & \textbf{0.694} & \textcolor{gray}{0.032} & \textcolor{gray}{0.992} & \textcolor{gray}{0.079} & \textbf{0.043} & \textbf{0.987} & 0.196 & 0.764 & 2.67 \\
MoGe~\citep{moge} & 0.281 & 0.627 & 0.132 & 0.896 & 0.126 & 0.086 & 0.967 & 0.101 & 0.971 & 2.25 \\
Ours & \textbf{0.257} & 0.617 & \textbf{0.100} & \textbf{0.961} & \textbf{0.082} & 0.068 & 0.979 & \textbf{0.091} & \textbf{0.976} & \textbf{1.25} \\
\bottomrule
\toprule
& $\text{Rel}^{d}\!\downarrow$ & $\delta^{d}\!\uparrow$ & $\text{Rel}^{d}\!\downarrow$ & $\delta^{d}\!\uparrow$ & $\text{TAE}^{d}\!\downarrow$ & $\text{Rel}^{d}\!\downarrow$ & $\delta^{d}\!\uparrow$ & $\text{Rel}^{d}\!\downarrow$ & $\delta^{d}\!\uparrow$ & Rank $\!\downarrow$ \\
\midrule
DepthPro~\citep{depthpro} & 0.363 & 0.476 & \textcolor{gray}{0.089} & \textcolor{gray}{0.929} & \textcolor{gray}{0.065} & 0.056 & 0.973 & 0.092 & 0.912 & 3.33 \\
VGGT~\citep{wang2025vggt} & 0.359 & \textbf{0.680} & \textcolor{gray}{0.029} & \textcolor{gray}{0.989} & \textcolor{gray}{0.048} & \textbf{0.040} & \textbf{0.981} & 0.187 & 0.728 & 2.67 \\
MoGe~\citep{moge} & 0.255 & 0.603 & 0.130 & 0.852 & 0.077 & 0.081 & 0.959 & 0.087 & 0.958 & 2.50 \\
Ours & \textbf{0.216} & 0.648 & \textbf{0.081} & \textbf{0.941} & \textbf{0.049} & 0.055 & 0.971 & \textbf{0.081} & \textbf{0.965} & \textbf{1.25} \\
\bottomrule
\end{tabular}
}
\end{table*}

\begin{table*}[t]
\caption{
  \textbf{Ablation and video-depth evaluations.}
  Left: position encoding methods on 270-frame sequences exceeding our 24-frame training sequences, where RoPE+ combines NTK-adapted rotary encoding with sequence stretching training.
  Right: video depth methods on 300 frames at 378\texttimes{}672 resolution with affine-invariant alignment.
}
\label{tab:ablate_positional}
\centering
\begin{minipage}{.58\textwidth}
    \centering
    \setlength{\tabcolsep}{2.5pt}
    \resizebox{\textwidth}!{
    \begin{tabular}{l|ccc|ccc|cc}
        \toprule
        \multirow{2}{*}{Pos. Encoding} & \multicolumn{3}{c|}{Sintel} & \multicolumn{3}{c|}{ScanNet} & \multicolumn{2}{c}{Bonn} \\
        & $\text{Rel}^{p}\!\downarrow$ & $\delta^{p}\!\uparrow$ & $\text{TAE}^{p}\!\downarrow$ & $\text{Rel}^{p}\!\downarrow$ & $\delta^{p}\!\uparrow$ & $\text{TAE}^{p}\!\downarrow$ & $\text{Rel}^{p}\!\downarrow$ & $\delta^{p}\!\uparrow$ \\
        \midrule
        None & 0.304 & \textbf{0.503} & 0.426 & 0.163 & 0.878 & 0.089 & 0.118 & 0.958 \\
        APE & 0.324 & 0.475 & 0.451 & 0.153 & 0.895 & 0.089 & 0.115 & 0.956 \\
        RoPE & 0.307 & 0.491 & 0.410 & 0.140 & 0.915 & 0.092 & 0.103 & \textbf{0.964} \\
        RoPE+ & \textbf{0.304} & \textbf{0.503} & \textbf{0.394} & \textbf{0.138} & \textbf{0.923} & \textbf{0.086} & \textbf{0.095} & 0.963 \\
        \midrule
        \multirow{2}{*}{Pos. Encoding} & \multicolumn{3}{c|}{Sintel} & \multicolumn{3}{c|}{ScanNet} & \multicolumn{2}{c}{Bonn} \\
        & $\text{Rel}^{d}\!\downarrow$ & $\delta^{d}\!\uparrow$ & $\text{TAE}^{d}\!\downarrow$ & $\text{Rel}^{d}\!\downarrow$ & $\delta^{d}\!\uparrow$ & $\text{TAE}^{d}\!\downarrow$ & $\text{Rel}^{d}\!\downarrow$ & $\delta^{d}\!\uparrow$ \\
        \midrule
        None & 0.261 & 0.547 & 0.246 & 0.107 & 0.896 & 0.053 & 0.073 & 0.954 \\
        APE & 0.279 & 0.526 & 0.252 & 0.108 & 0.889 & 0.052 & 0.076 & 0.947 \\
        RoPE & 0.261 & 0.547 & 0.235 & 0.097 & 0.911 & 0.055 & \textbf{0.062} & \textbf{0.959} \\
        RoPE+ & \textbf{0.253} & \textbf{0.563} & \textbf{0.222} & \textbf{0.095} & \textbf{0.919} & \textbf{0.047} & 0.064 & \textbf{0.959} \\
        \bottomrule
    \end{tabular}}
\end{minipage}%
\hfill
\begin{minipage}{.40\textwidth}
    \centering
    \setlength{\tabcolsep}{2pt}
    \resizebox{\textwidth}!{
    \begin{tabular}{l|cc|cc|c}
        \toprule
        \multirow{2}{*}{Method} & \multicolumn{2}{c|}{Sintel} & \multicolumn{2}{c|}{Bonn} & \multirow{2}{*}{FPS} \\
        & $\text{Rel}^{d}\!\downarrow$ & $\delta^{d}\!\uparrow$ & $\text{Rel}^{d}\!\downarrow$ & $\delta^{d}\!\uparrow$ & \\
        \midrule
        DepthCrafter & 0.30 & 0.70 & 0.13 & 0.85 & 0.94 \\
        Video Depth Any. & 0.30 & 0.64 & 0.07 & 0.96 & 4.47 \\
        DepthAnyVideo & 0.41 & 0.66 & \textbf{0.06} & \textbf{0.97} & 6.48 \\
        \textbf{\model{} (Ours)} & \textbf{0.20} & \textbf{0.73} & \textbf{0.06} & \textbf{0.97} & \textbf{39.1} \\
        \midrule
        \multirow{2}{*}{Method} & \multicolumn{2}{c|}{ScanNet} & \multicolumn{2}{c|}{KITTI} & \multirow{2}{*}{Time (s)} \\
        & $\text{Rel}^{d}\!\downarrow$ & $\delta^{d}\!\uparrow$ & $\text{Rel}^{d}\!\downarrow$ & $\delta^{d}\!\uparrow$ & \\
        \midrule
        DepthCrafter & 0.17 & 0.73 & 0.15 & 0.77 & 320.1 \\
        Video Depth Any. & \textbf{0.09} & 0.92 & 0.08 & 0.95 & 67.2 \\
        DepthAnyVideo & \textbf{0.09} & \textbf{0.93} & 0.11 & 0.89 & 46.3 \\
        \textbf{\model{} (Ours)} & \textbf{0.09} & \textbf{0.93} & \textbf{0.07} & \textbf{0.97} & \textbf{7.7} \\
        \bottomrule
    \end{tabular}}
\end{minipage}
\end{table*}

\begin{figure*}[p]
    \centering
    \setlength{\tabcolsep}{1pt}
    \begin{tabular}{cccc}
        \includegraphics[width=0.24\textwidth]{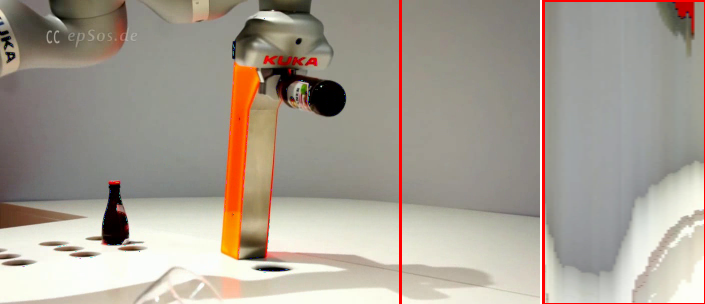} &
        \includegraphics[width=0.24\textwidth]{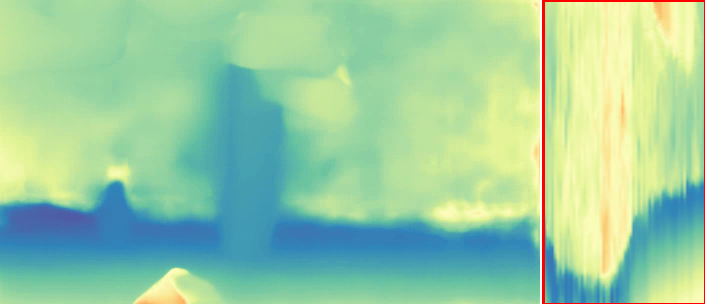} &
        \includegraphics[width=0.24\textwidth]{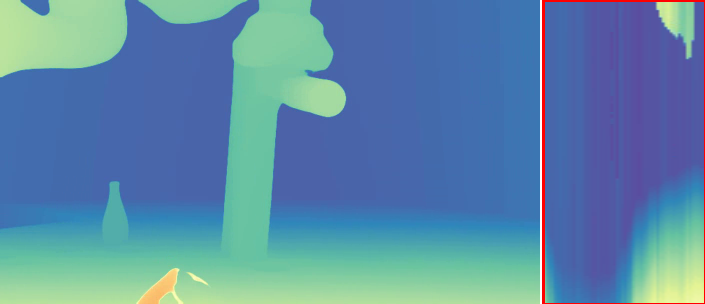} &
        \includegraphics[width=0.24\textwidth]{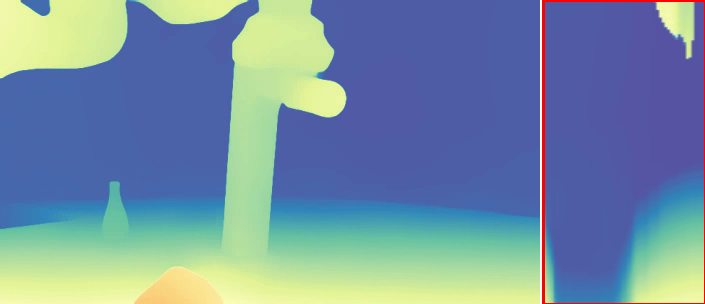} \\[3pt]

        \includegraphics[width=0.24\textwidth]{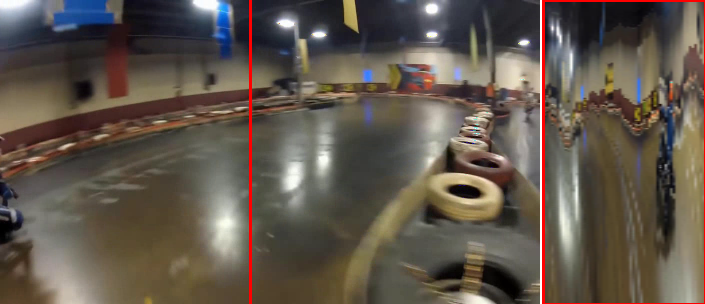} &
        \includegraphics[width=0.24\textwidth]{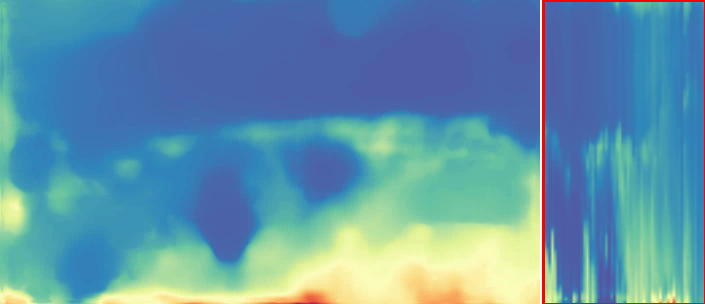} &
        \includegraphics[width=0.24\textwidth]{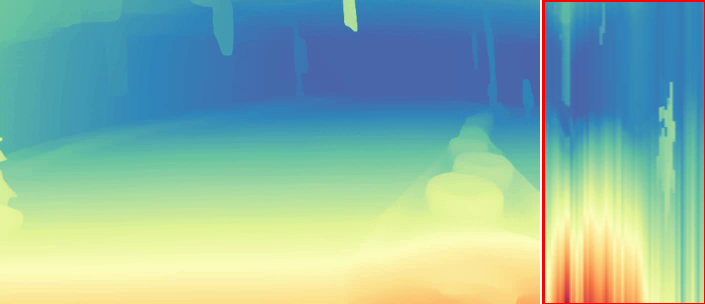} &
        \includegraphics[width=0.24\textwidth]{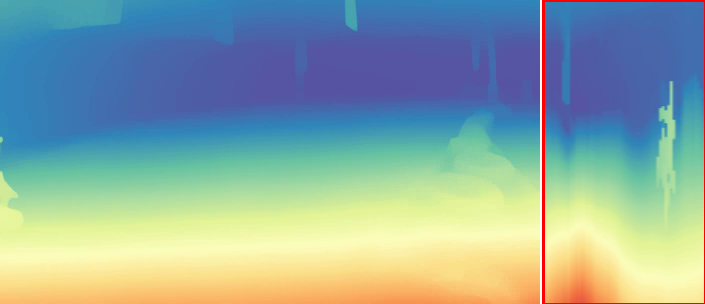} \\[3pt]

        \includegraphics[width=0.24\textwidth]{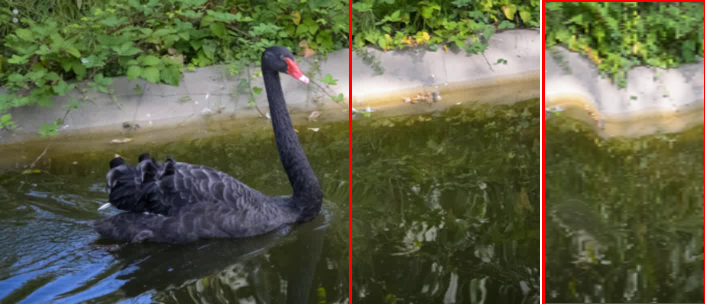} &
        \includegraphics[width=0.24\textwidth]{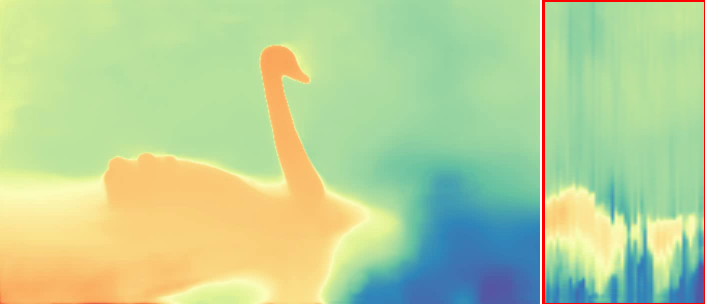} &
        \includegraphics[width=0.24\textwidth]{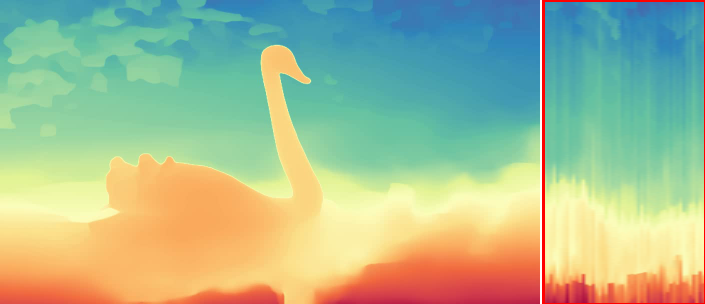} &
        \includegraphics[width=0.24\textwidth]{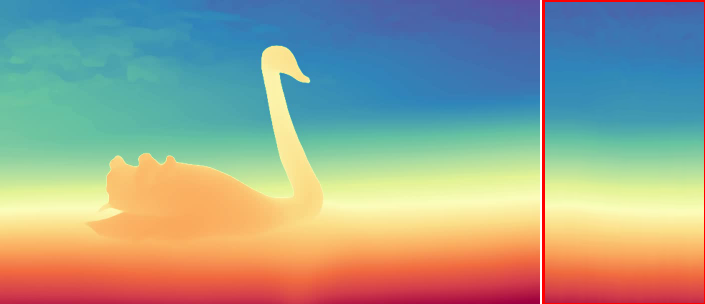} \\[3pt]

        \includegraphics[width=0.24\textwidth]{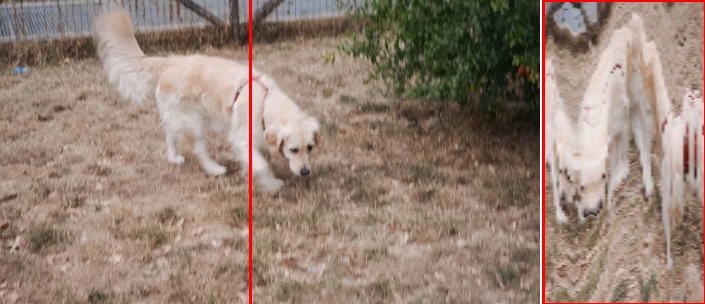} &
        \includegraphics[width=0.24\textwidth]{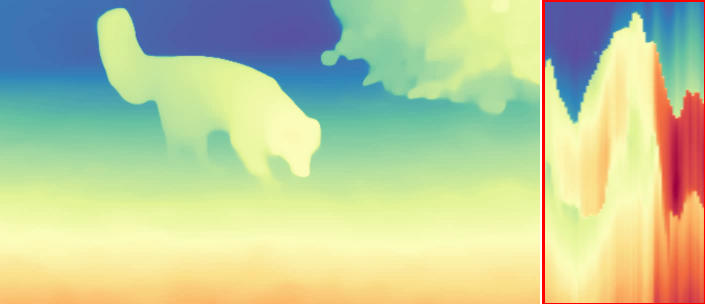} &
        \includegraphics[width=0.24\textwidth]{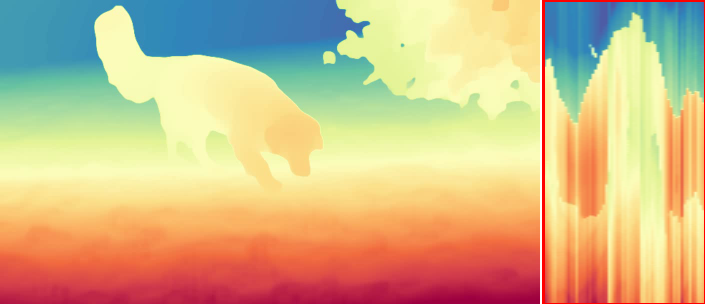} &
        \includegraphics[width=0.24\textwidth]{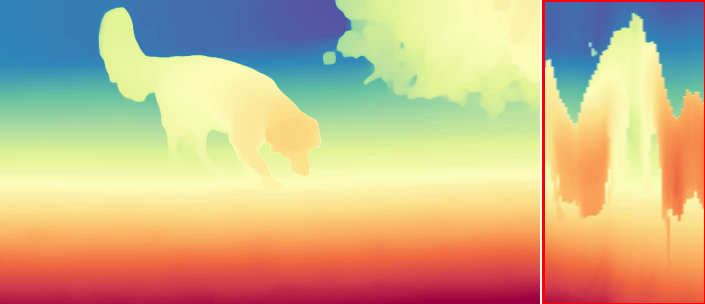} \\[3pt]

        \includegraphics[width=0.24\textwidth]{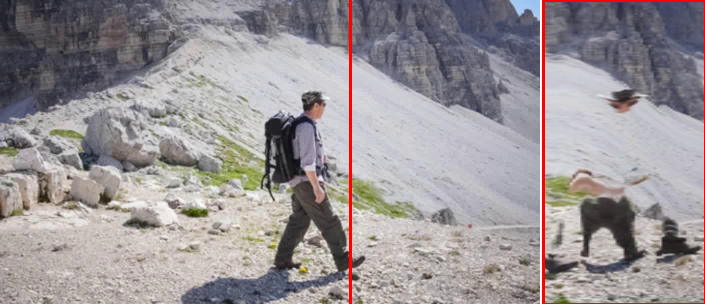} &
        \includegraphics[width=0.24\textwidth]{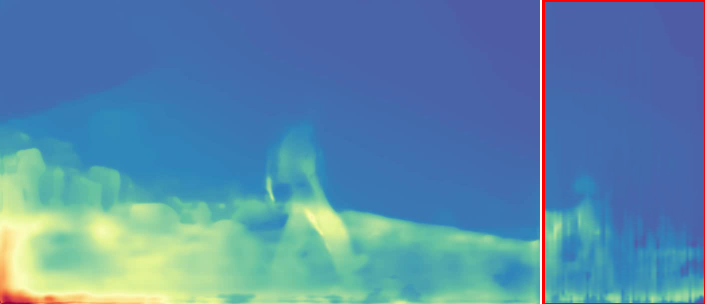} &
        \includegraphics[width=0.24\textwidth]{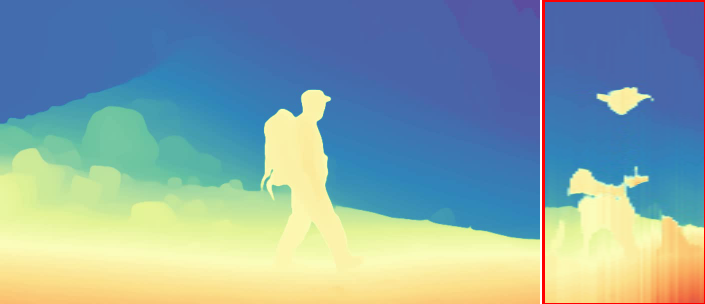} &
        \includegraphics[width=0.24\textwidth]{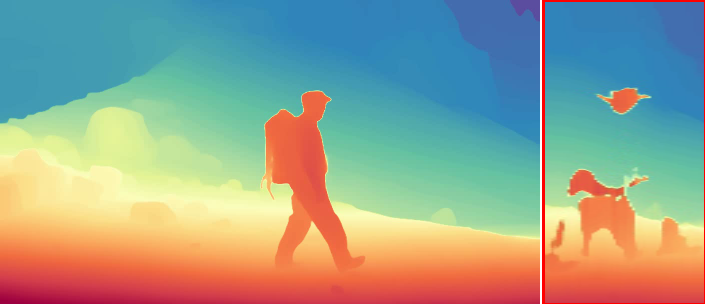} \\[3pt]

        \includegraphics[width=0.24\textwidth]{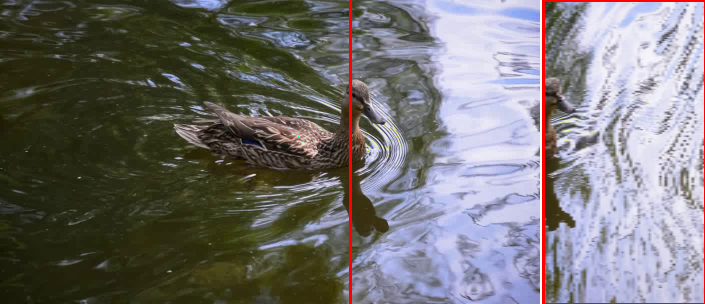} &
        \includegraphics[width=0.24\textwidth]{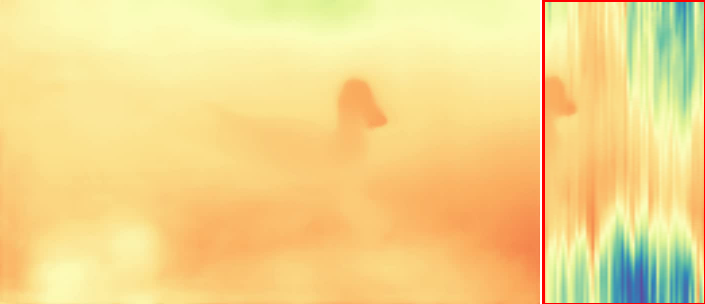} &
        \includegraphics[width=0.24\textwidth]{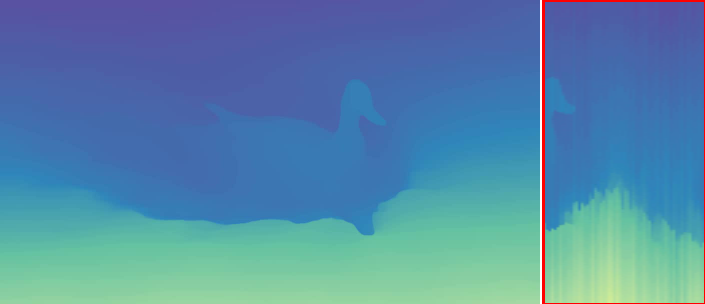} &
        \includegraphics[width=0.24\textwidth]{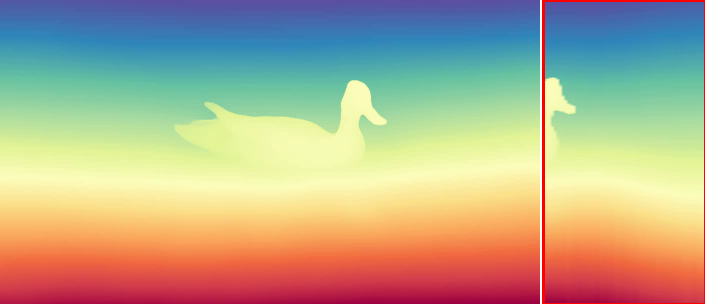} \\[3pt]

        \includegraphics[width=0.24\textwidth]{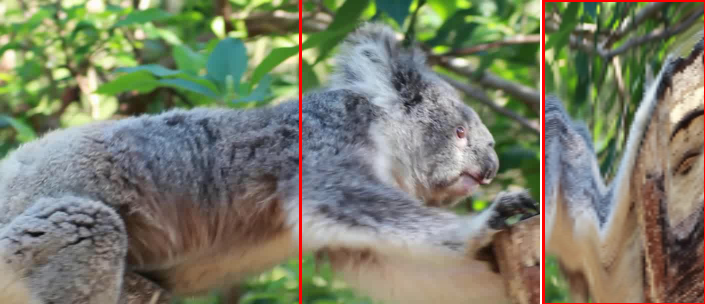} &
        \includegraphics[width=0.24\textwidth]{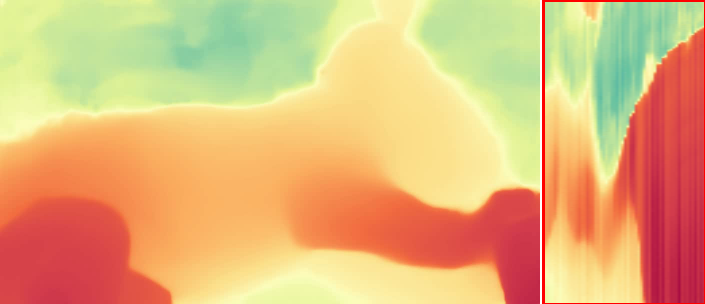} &
        \includegraphics[width=0.24\textwidth]{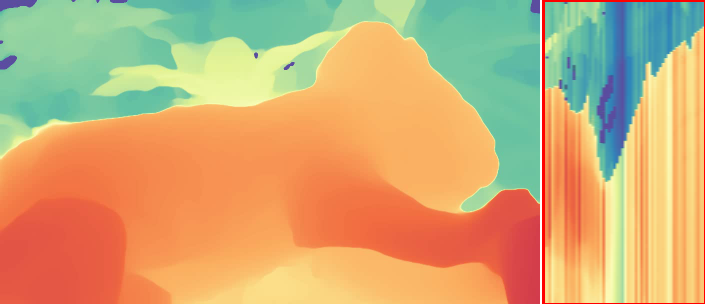} &
        \includegraphics[width=0.24\textwidth]{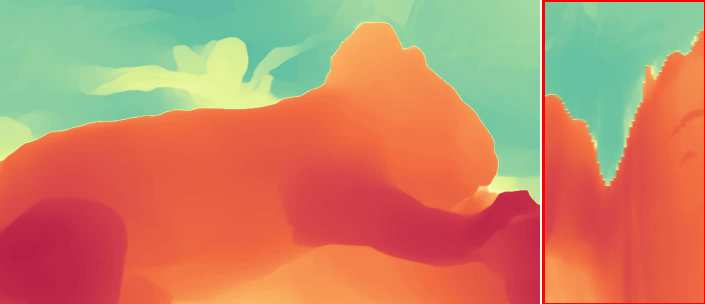} \\[3pt]

        \includegraphics[width=0.24\textwidth]{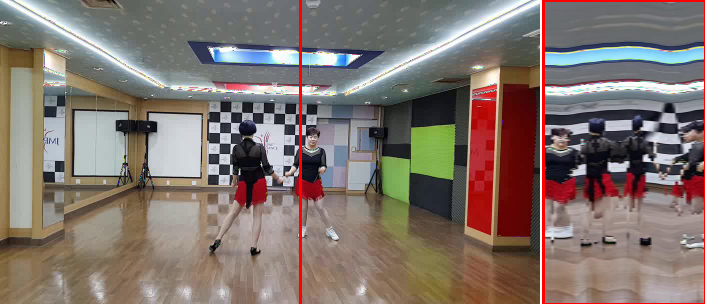} &
        \includegraphics[width=0.24\textwidth]{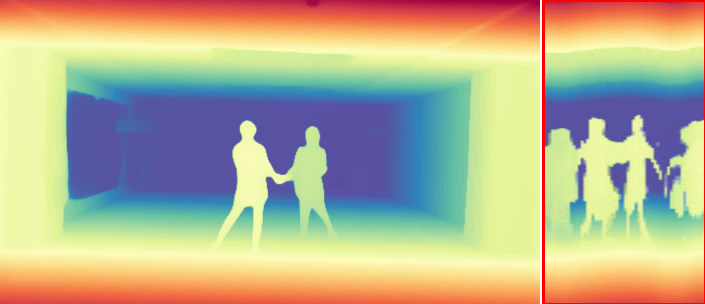} &
        \includegraphics[width=0.24\textwidth]{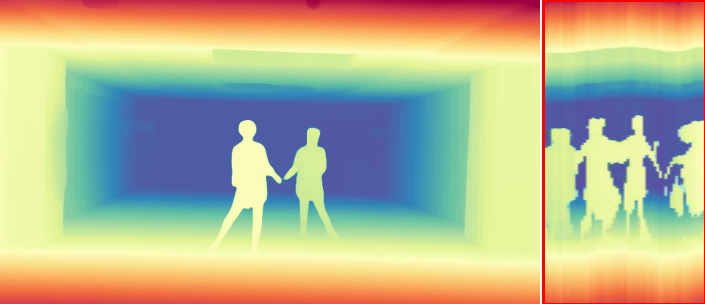} &
        \includegraphics[width=0.24\textwidth]{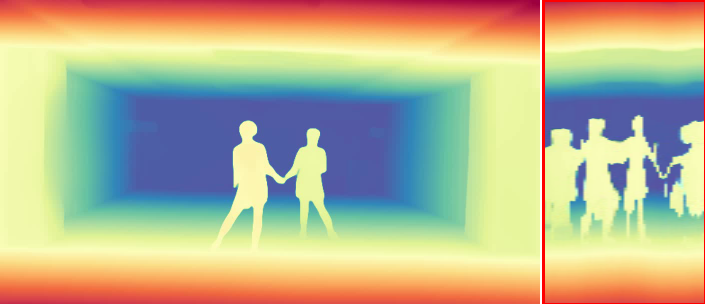} \\[3pt]

        \includegraphics[width=0.24\textwidth]{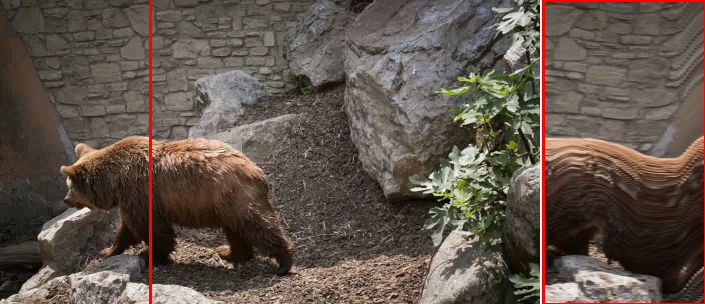} &
        \includegraphics[width=0.24\textwidth]{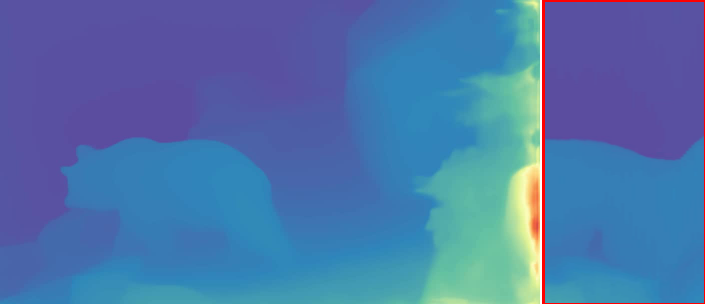} &
        \includegraphics[width=0.24\textwidth]{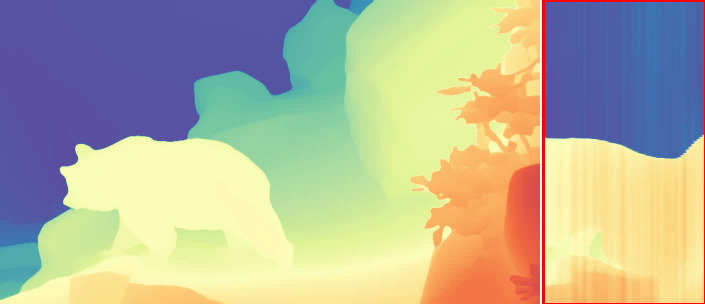} &
        \includegraphics[width=0.24\textwidth]{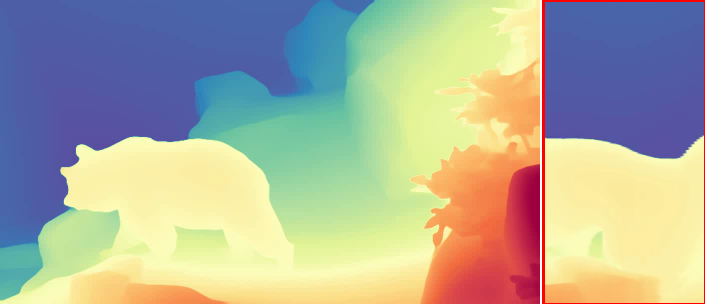} \\[3pt]

        \multicolumn{1}{c}{\textbf{Input Video}} &
        \multicolumn{1}{c}{\textbf{VGGT}} &
        \multicolumn{1}{c}{\textbf{MoGe}} &
        \multicolumn{1}{c}{\textbf{Ours}} \\
    \end{tabular}
    \caption{\textbf{Qualitative visualizations of depth predictions across diverse scenarios.} Each row shows an input frame from DAVIS~\citep{davis} with its corresponding spacetime slice (right portion), comparing depth predictions from VGGT, MoGe, and our method.}
    \Description{Grid of several video examples comparing input frames and corresponding depth spacetime-slice visualizations for VGGT, MoGe, and \model.}
    \label{fig:qualitative_comparison}
\end{figure*}

\begin{table*}[t]
\caption{
  \textbf{Effectiveness of single-pass processing for long sequences.}
  We compare directly processing entire 270-frame sequences with our frequency-modulated position encoding (Single-Pass) against traditional sliding window approach with overlapping frames.
}
\label{tab:ablate_infer}
\centering
\resizebox{\textwidth}{!}{%
\begin{tabular}{l|cccc|cccc|cccc}
\toprule
\multirow{2}{*}{Inference Method} & \multicolumn{4}{c|}{ScanNet (270 frames)} & \multicolumn{4}{c|}{KITTI (270 frames)} & \multicolumn{4}{c}{DDAD} \\
& $\text{Rel}^{p}\!\downarrow$ & $\delta^{p}\!\uparrow$ & $\text{Rel}^{d}\!\downarrow$ & $\delta^{d}\!\uparrow$ & $\text{Rel}^{p}\!\downarrow$ & $\delta^{p}\!\uparrow$ & $\text{Rel}^{d}\!\downarrow$ & $\delta^{d}\!\uparrow$ & $\text{Rel}^{p}\!\downarrow$ & $\delta^{p}\!\uparrow$ & $\text{Rel}^{d}\!\downarrow$ & $\delta^{d}\!\uparrow$ \\
\midrule
Sliding Window & 0.114 & 0.935 & 0.098 & 0.908 & 0.102 & 0.963 & 0.097 & 0.930 & 0.192 & 0.863 & 0.115 & 0.894 \\
Single-Pass & \textbf{0.113} & \textbf{0.937} & \textbf{0.094} & \textbf{0.913} & \textbf{0.092} & \textbf{0.974} & \textbf{0.084} & \textbf{0.963} & \textbf{0.187} & \textbf{0.879} & \textbf{0.108} & \textbf{0.916} \\
\bottomrule
\end{tabular} 
}
\end{table*}

\begin{figure*}[t]
  \centering
  \includegraphics[width=0.9\textwidth]{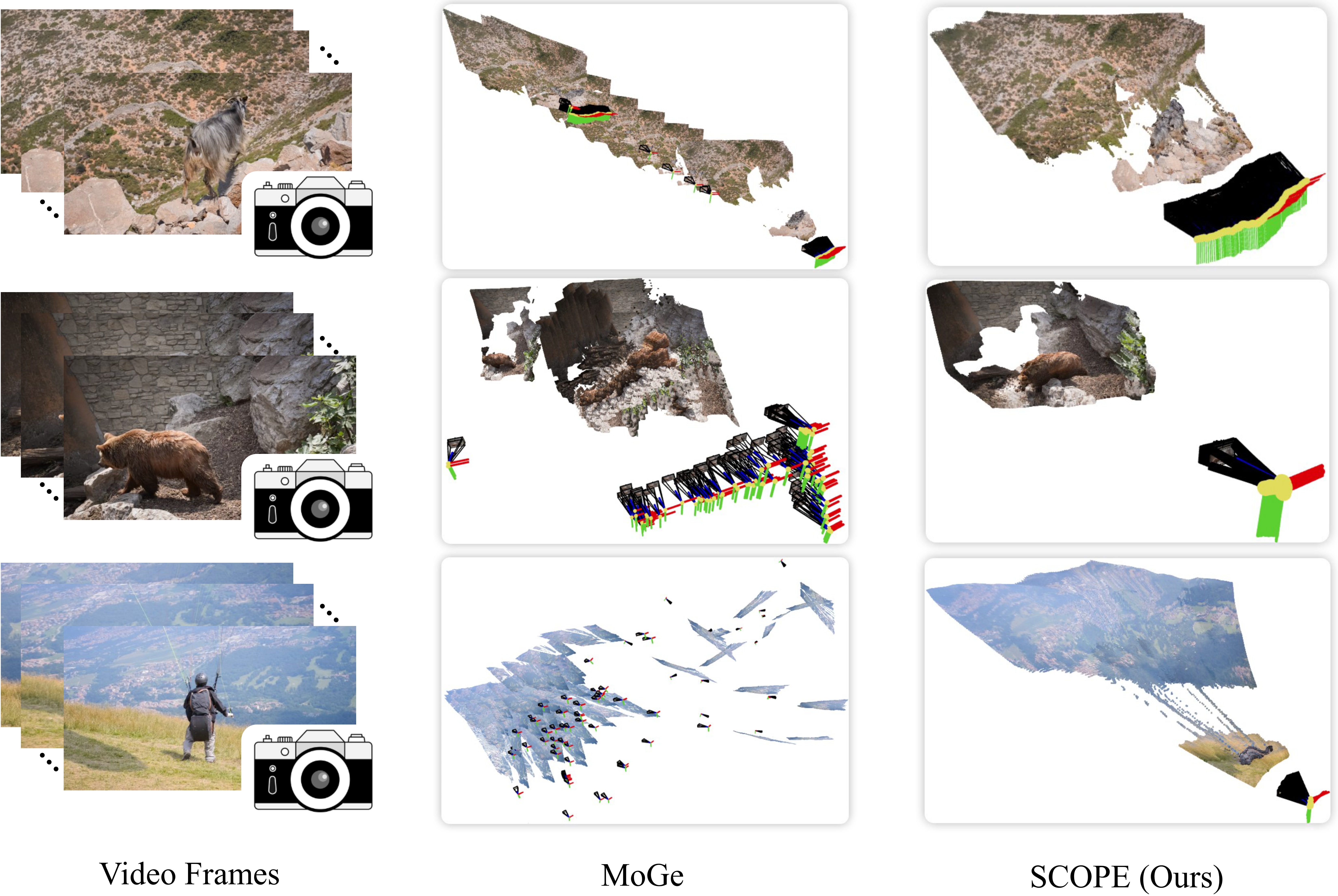}
  \caption{
       \textbf{4D reconstruction comparison using MegaSAM~\citep{megasam}.} Our method enables coherent multi-view reconstruction from video sequences (right), while MoGe (middle) produces fragmented results with significant distortion. Input video frames shown on left.
  }
  \Description{Qualitative comparison of 4D reconstructions from MegaSAM using point maps predicted by MoGe versus \model, with input frames shown for reference.}
  \label{fig:megasam}
\end{figure*}

\subsection{Evaluation}
\paragraph{Evaluation datasets.} We evaluate on five diverse datasets spanning various scenarios:
\textbf{Sintel}~\citep{sintel} consists of 23 synthetic videos with 50 frames each, providing precise depth labels in complex scenes with challenging lighting and motion.
\textbf{ScanNet v2}~\citep{scannet} includes 100 indoor test videos with rich geometric structures, from which we sample every third frame to create 90-frame sequences for standard evaluation.
\textbf{Bonn}~\citep{bonn} contains 26 dynamic videos with prominent foreground motions, where we use frames 30-140 to assess robustness to object movement.
\textbf{KITTI}~\citep{kitti} provides 13 outdoor driving sequences, from which we use the first 110 frames per sequence from the full validation set to evaluate performance in structured environments.
\textbf{DDAD}~\citep{ddad} is an autonomous driving dataset featuring diverse outdoor scenes captured across varying weather conditions and environments, with sequences ranging from 50 to 100 frames.
For ablation studies focusing on long-range temporal consistency, we extend our evaluation to 270-frame sequences using consistent sampling strategies across datasets where ground truth is available.
We also evaluate on 7-Scenes~\citep{7scenes} and TUM RGB-D sequences with the same global scale alignment and temporal alignment evaluation used throughout this section.

\paragraph{Quantitative results.} Table~\ref{tab:comp_point} presents our method's performance compared to state-of-the-art approaches. We report point map ($\text{Rel}^{p}$, $\delta^{p}$) and depth ($\text{Rel}^{d}$, $\delta^{d}$) metrics, along with temporal alignment error (TAE)~\citep{yang2024depthanyvideo}. All evaluations use a shared scale factor across entire sequences to assess global consistency.

\begin{sloppypar}
Our approach significantly outperforms previous methods, achieving the lowest average rank across all datasets. We achieve substantial improvements: 8.5\% $\text{Rel}^{p}$ reduction on Sintel, 24.2\% accuracy and 34.9\% temporal consistency ($\text{TAE}^{p}$) improvements on ScanNet, and 9.9\% $\text{Rel}^{p}$ reduction on KITTI compared to MoGe. Together with Table~\ref{tab:comp_point}, Table~\ref{tab:strong_baselines_compact} compares \model{} with metric-depth methods such as MoGe-2~\citep{moge2}, VGGT~\citep{wang2025vggt} and its derived $\pi^3$ model~\citep{pi3}, and DUSt3R-derived multiview methods~\citep{dust3r}, including MASt3R-SLAM~\citep{mast3rslam} and MonST3R~\citep{monst3r}. Although initialized from the original MoGe~\citep{moge} weights, \model{} achieves comparable point-map accuracy across these comparisons and gives the best or tied-best temporal alignment errors on all evaluated datasets, reducing both $\text{TAE}^{p}$ and $\text{TAE}^{d}$ relative to MoGe-2. It also remains competitive with $\pi^3$, MASt3R-SLAM, and MonST3R; compared with $\pi^3$, \model{} uses 322M rather than 959M parameters. Depth metrics show similar trends across datasets.
\end{sloppypar}

\begin{table}[t]
\caption{
  Comparison with other geometry methods on the evaluated datasets. Each entry reports $\delta^{p}\!\uparrow / \text{TAE}^{p}\!\downarrow / \text{TAE}^{d}\!\downarrow$.
}
\label{tab:strong_baselines_compact}
\centering
\begingroup
\setlength{\tabcolsep}{1.8pt}
\renewcommand{\arraystretch}{1.05}
\scriptsize
\resizebox{\columnwidth}{!}{%
\begin{tabular}{lcccc}
\toprule
Method & Sintel & Bonn & KITTI & TUM \\
\midrule
MoGe-2 & 0.604/0.450/0.284 & \textbf{0.982}/0.034/0.032 & \textbf{0.984}/0.068/0.037 & 0.954/0.059/0.051 \\
\makecell[l]{MASt3R-\\SLAM} & 0.553/0.395/0.229 & 0.936/0.031/0.029 & 0.621/0.117/0.086 & 0.914/0.058/0.049 \\
MonST3R & 0.553/0.503/0.284 & \textbf{0.982}/0.026/0.023 & 0.600/0.074/0.036 & 0.957/\textbf{0.042}/\textbf{0.032} \\
$\pi^3$ & \textbf{0.762}/0.649/0.417 & 0.959/\textbf{0.020}/\textbf{0.017} & 0.897/0.067/0.029 & \textbf{0.978}/0.043/\textbf{0.032} \\
\model{} & 0.635/\textbf{0.363}/\textbf{0.212} & 0.978/\textbf{0.020}/\textbf{0.017} & 0.962/\textbf{0.061}/\textbf{0.022} & 0.971/\textbf{0.042}/\textbf{0.032} \\
\bottomrule
\end{tabular}
}
\endgroup
\end{table}

The right panel of Table~\ref{tab:ablate_positional} evaluates our method against state-of-the-art video depth estimation approaches. Our method achieves competitive or superior accuracy across all datasets while demonstrating remarkable computational efficiency, processing sequences 6--42\texttimes{} faster than existing methods.

\paragraph{Qualitative comparison.} Figure~\ref{fig:qualitative_comparison} presents spacetime slice visualizations showing that our method maintains superior temporal consistency across diverse scenarios compared to VGGT and MoGe. Figure~\ref{fig:megasam} demonstrates the downstream impact, with our approach enabling coherent 4D reconstructions via MegaSAM~\citep{megasam} while MoGe produces fragmented results under identical conditions. To quantify this reconstruction quality, we evaluated camera pose accuracy using our predicted point maps as input to MegaSAM on the Sintel dataset. Our method achieves significant improvements with ATE of 0.035 and RTE of 0.014, outperforming both MonST3R (ATE: 0.078, RTE: 0.038) and MoGe (ATE: 0.087, RTE: 0.033) by 55\% and 60\% respectively in ATE, and 63\% and 58\% respectively in RTE. Notably, our method achieved 100\% success rate while MoGe failed completely on 2 scenes. We further evaluate on the 7-Scenes dataset~\citep{7scenes}, where our method achieves an ATE of 0.044m via MegaSAM, outperforming MASt3R-SLAM~\citep{mast3rslam} (0.047m) by 7.2\%.

\subsection{Ablation Study}

\paragraph{Extrapolation strategies for long sequences.} The left panel of Table~\ref{tab:ablate_positional} compares position encoding strategies under long-sequence evaluation, using full Sintel clips together with 270-frame ScanNet and Bonn sequences: no encoding (None), absolute (APE), RoPE with NTK adaptation, and RoPE+ adding sequence stretching during training.
Across these settings, RoPE+ provides the strongest overall trade-off. Compared to using no position encoding, it reduces $\text{TAE}^{p}$ from 0.426 to 0.394 on Sintel and from 0.089 to 0.086 on ScanNet, while improving $\delta^{p}$ from 0.878 to 0.923 on ScanNet. The same trend holds on shorter sequences. On 90-frame ScanNet, RoPE+ improves $\text{Rel}^{p}$ from 0.145 to 0.125, $\delta^{p}$ from 0.92 to 0.95, $\text{TAE}^{p}$ from 0.081 to 0.078, and $\text{TAE}^{d}$ from 0.048 to 0.045 over the no-encoding baseline. On the standard Bonn evaluation clips, it reduces $\text{Rel}^{p}$ from 0.101 to 0.079 and $\text{Rel}^{d}$ from 0.061 to 0.050. RoPE+ is best or tied best on every reported short-sequence metric. The substantial improvements from our sequence stretching training technique (RoPE+ vs. standard RoPE) confirm that explicitly simulating extrapolation during training significantly enhances the model's ability to handle sequences far exceeding the training context window.

\begin{table}[t]
\caption{
  Component ablations on ScanNet-270 for temporal modules, loss terms, and appearance augmentation. Each switch is evaluated independently with the same backbone and learning schedule, so single-component rows are not strictly cumulative.
}
\label{tab:component_ablations}
\centering
\begingroup
\renewcommand{\arraystretch}{1.08}
\setlength{\tabcolsep}{4pt}
\small
\resizebox{.96\columnwidth}{!}{%
\begin{tabular}{p{1.15cm}p{2.35cm}ccc}
\toprule
Study & Setting & $\text{Rel}^{p}\!\downarrow$ & $\text{TAE}^{p}\!\downarrow$ & $\text{TAE}^{d}\!\downarrow$ \\
\midrule
Temporal & MoGe & 0.168 & 0.121 & 0.068 \\
& \makecell[l]{+ temporal\\modules} & \textbf{0.125} & \textbf{0.097} & \textbf{0.062} \\
\midrule
Loss & base & \textbf{0.1245} & 0.0974 & 0.0616 \\
& + $\mathcal{L}_{cross}$ & 0.1246 & 0.0945 & 0.0587 \\
& + $\mathcal{L}_{temp}$ & 0.1968 & 0.0882 & 0.0479 \\
& + both & 0.1376 & \textbf{0.0860} & \textbf{0.0474} \\
\midrule
Appear. & w/o aug. & \textbf{0.1114} & 0.1003 & 0.0650 \\
& with aug. & 0.1245 & \textbf{0.0974} & \textbf{0.0616} \\
\bottomrule
\end{tabular}
}
\endgroup
\end{table}

\begin{sloppypar}
\paragraph{Effectiveness of temporal and geometric constraints.} We evaluate our hierarchical temporal supervision ($\mathcal{L}_{temp}$) and cross-frame geometric constraints ($\mathcal{L}_{cross}$) on Sintel, ScanNet, and DDAD datasets. The combination achieves pointmap temporal consistency improvements of 9.53\% on average and depth temporal consistency improvements of 18.4\% compared to baseline MoGe constraints. Table~\ref{tab:component_ablations} reports component-wise studies on ScanNet-270. Adding the temporal modules improves $\text{Rel}^{p}/\text{TAE}^{p}/\text{TAE}^{d}$ from 0.168/0.121/0.068 to 0.125/0.097/0.062. In the loss isolation study, $\mathcal{L}_{cross}$ reduces $\text{TAE}^{p}/\text{TAE}^{d}$ from 0.0974/0.0616 to 0.0945/0.0587, $\mathcal{L}_{temp}$ reduces them to 0.0882/0.0479, and combining both yields the best temporal consistency, 0.0860/0.0474. Appearance augmentation trades a small amount of single-frame accuracy for lower temporal error, reducing $\text{TAE}^{p}$ from 0.1003 to 0.0974 and $\text{TAE}^{d}$ from 0.0650 to 0.0616.
\end{sloppypar}

\paragraph{Single-pass vs. sliding window inference.} 
Table~\ref{tab:ablate_infer} compares our single-pass processing approach with traditional sliding window techniques~\citep{video_depth_anything} for handling long sequences. Our method directly processes entire 270-frame sequences in a single forward pass.
This approach not only eliminates computational redundancy but also consistently improves performance across all datasets. On KITTI, single-pass processing reduces $\text{Rel}^{p}$ by 9.8\% compared to sliding window approaches, highlighting the benefits of maintaining global context across the entire sequence rather than processing overlapping segments independently.

\paragraph{Computational efficiency.} On an NVIDIA H20 GPU with FP16 inference, our model processes 300 frames at 378\texttimes{}672 in 7.68 seconds (39.1 FPS), using 76.53 GB memory. The optimization takes 0.337 seconds in total, or 1.12 ms per frame.

\section{Conclusion}

We introduced \model{} for monocular video geometry estimation, targeting high geometric accuracy and long-range temporal consistency in a single pass. It combines viewpoint-invariant geometry, appearance-invariant learning, and frequency-modulated temporal positioning to produce scale-invariant point maps over long sequences. Experiments show consistent gains across diverse datasets while preserving fine-grained details and global structure. A remaining limitation is the reliance on MegaSAM~\citep{megasam} for camera poses; future work will estimate camera trajectories directly within the network.

\begin{acks}
This work is supported by the National Natural Science Foundation of China (No. 62422606) and Hong Kong Research Grants Council General Research Fund (No. 17213925).
\end{acks}

\bibliographystyle{ACM-Reference-Format}
\bibliography{reference}

\end{document}